\documentclass[letterpaper, 10 pt, conference]{ieeeconf}  %

\IEEEoverridecommandlockouts                              %

\usepackage{amsmath} %
\usepackage{amsfonts} %
\usepackage{amssymb}  %
\usepackage{graphicx}
\usepackage{hyperref}
\usepackage{mathtools}
\usepackage{comment}

\usepackage{filecontents}
\usepackage{xcolor}
\usepackage{graphicx, subcaption}
\usepackage[linesnumbered, ruled,vlined]{algorithm2e}
\usepackage{comment}

\usepackage[font=footnotesize,labelfont=footnotesize]{subcaption}
\usepackage[font=footnotesize,labelfont=footnotesize]{caption}

\SetKwInput{KwInput}{Input}                %
\SetKwInput{KwOutput}{Output}              %

\usepackage{amssymb}
\usepackage{amsmath, bm}
\usepackage{verbatim}

\usepackage{multirow}
\usepackage{hhline}
\renewcommand{\vec}{\mathbf}

\usepackage{cite}

\usepackage{color}

\usepackage[normalem]{ulem}

\usepackage{tabularx}

\renewcommand{\vec}{\mathbf}

\newcommand\Reals{\mathbb{R}}

\DeclareMathOperator*{\argmax}{arg\,max}

\definecolor{purple}{RGB}{210, 0, 210} %
\definecolor{international_orange}{RGB}{240, 74, 0}

\graphicspath{{figures/}}

\definecolor{D}{RGB}{196, 114, 0}
\definecolor{Q}{RGB}{18, 113, 148}

\title{\LARGE \bf
Planning Sensing Sequences for Subsurface 3D Tumor Mapping}

\author{Brian Y. Cho, Tucker Hermans, and Alan Kuntz%
\thanks{The authors are with the Robotics Center and School of Computing,
        University of Utah, Salt Lake City, UT 84112, USA. TH is also affiliated with NVIDIA.
        {\tt\small \{brian.cho,tucker.hermans,alan.kuntz\}@utah.edu}}%
}

\usepackage{color}

\begin{document}

\maketitle

\begin{abstract}
Surgical automation has the potential to enable increased precision and reduce the per-patient workload of overburdened human surgeons.
An effective automation system must be able to sense and map subsurface anatomy, such as tumors, efficiently and accurately.
In this work, we present a method that plans a sequence of sensing actions to map the 3D geometry of subsurface tumors.
We leverage a sequential Bayesian Hilbert map to create a 3D probabilistic occupancy model that represents the likelihood that any given point in the anatomy is occupied by a tumor, conditioned on sensor readings.
We iteratively update the map, utilizing Bayesian optimization to determine sensing poses that explore unsensed regions of anatomy and exploit the knowledge gained by previous sensing actions.
We demonstrate our method's efficiency and accuracy in three anatomical scenarios including a liver tumor scenario generated from a real patient's CT scan. 
The results show that our proposed method significantly outperforms comparison methods in terms of efficiency while detecting subsurface tumors with high accuracy.

\end{abstract}

\section{Introduction}

Surgical automation~\cite{Yang2018_SR,Yip2018_EMR} has the potential to increase precision and reduce the per-patient workload of surgeons, a group already stretched thin by a general population that is rapidly outgrowing surgical resources~\cite{Moffatt2018_JTCS,Ellison2020_Surgery}.
An effective autonomous surgical system must be able to sense, map, and reason about a patient's anatomy below the visible surface of organs.

\begin{figure}[t!]
    \centering
   \begin{subfigure}{0.22\textwidth}
	\includegraphics[width=\textwidth]{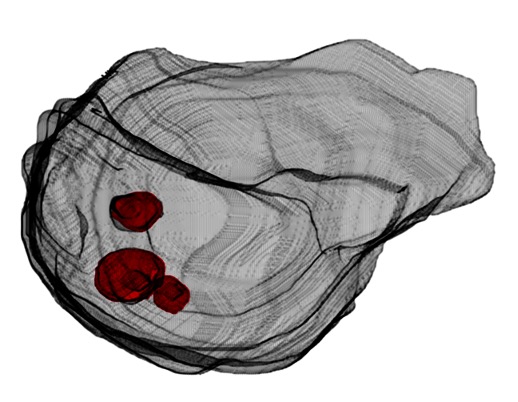}
    \caption{Segmentation side view}
    \label{fig:LiTS_Sideview}
  \end{subfigure}
  \begin{subfigure}{0.22\textwidth}
  \captionsetup{skip=-1pt}
	\includegraphics[width=\textwidth]{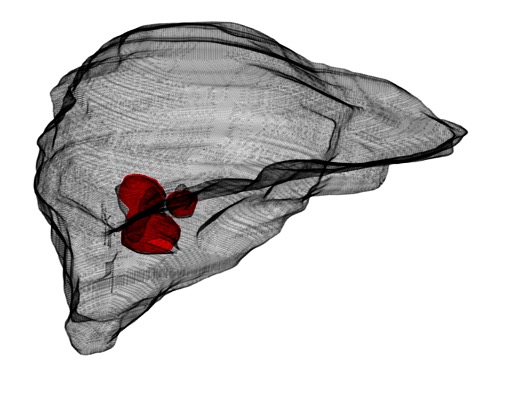}
    \caption{Segmentation top view}
    \label{fig:LiTS_Topview}
  \end{subfigure}
  \begin{subfigure}{0.22\textwidth}
	\includegraphics[width=\textwidth]{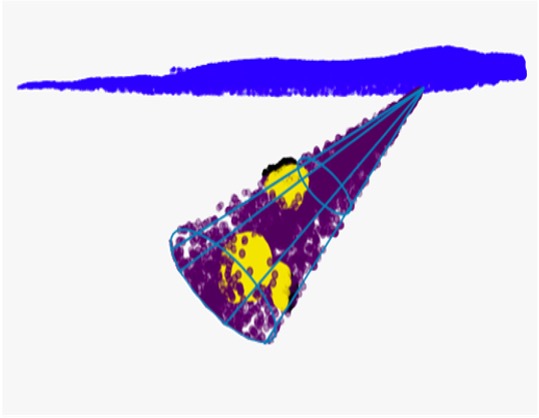}
    \caption{Sensing action side view}
    \label{fig:data_Sideview}
  \end{subfigure}
  \begin{subfigure}{0.22\textwidth}
	\includegraphics[width=\textwidth]{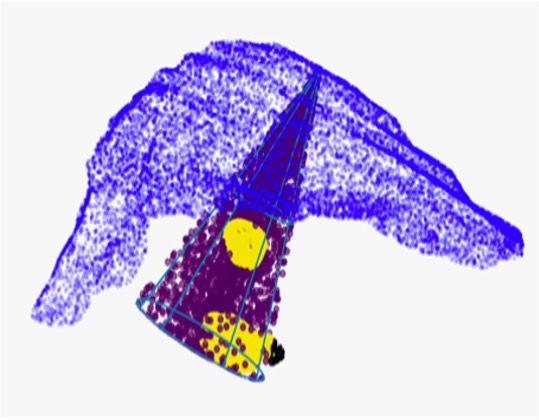}
    \caption{Sensing action top view}
    \label{fig:data_Topview}
  \end{subfigure}
  \caption{Example tumor localization scenario. (a,b) A liver (black) and 3 tumors (red) segmented from a CT scan in the liver tumor segmentation (LiTS) dataset. (c,d) A cone-shaped sensing volume (purple) normal to the liver surface (blue) detects the embedded tumors (yellow).
  }
  \label{fig:intro}
\vspace{-15pt}
\end{figure}

Consider the case of resecting subsurface tumors in an organ.
It is imperative to have an accurate understanding of the location and geometry of the tumors in the organ prior to resection in order to minimize the damage to healthy tissue while ensuring all cancerous tissue is safely removed (see Fig.~\ref{fig:intro}).
While pre-operative imaging techniques such as computed tomography (CT) can provide general knowledge of the anatomy, it may change before or during surgery.
This necessitates the use of intraoperative sensing and mapping of the tumors, e.g., through the use of an ultrasound probe.
Such sensing should be both accurate and efficient in order to reduce the overall time required for the surgical procedure.
In this work, we present a method to enable an autonomous surgical system to accurately and efficiently map subsurface patient anatomy, such as tumors inside an organ.

Specifically, we present a method that utilizes a probabilistic model of anatomical geometry to iteratively determine sensing actions that improve the model's understanding of the geometry.
To do so, we wrap a Bayesian optimization~\cite{Pelikan1999_GECCO} framework around a probabilistic representation of 3D geometry, called a Bayesian Hilbert map~\cite{Ramos2016_IJRR,senanayake2017bayesian}, that utilizes sensor information to determine the likelihood that any particular point in the organ is occupied by a tumor.
Bayesian Hilbert maps enable us to iteratively update the occupancy probability of a given continuous anatomical environment as we obtain sensor readings.
The Bayesian Hilbert map then serves as a continuous posterior distribution on tumor occupancy at each iteration, enabling the Bayesian optimization to determine the next sensing action.
Our method utilizes an objective function in the optimization that balances the exploration of unsensed regions of anatomy while reducing the uncertainty around regions where tumors have already been identified.
In this way the method is able to accurately map the geometry of the subsurface tumors in a relatively few number of sensing steps.

While prior work has focused on sensing and mapping in 2D~\cite{garg2016tumor, nichols2015methods}, our use of Bayesian Hilbert maps enables reasoning over 3D geometry accurately and efficiently.
We evaluate the performance of our method in multiple surgery-inspired scenarios, including environments in which tumors are randomly placed, evaluating generalizability, as well as a real-life scenario generated via the liver tumor segmentation (LiTS) dataset~\cite{bilic2019liver}.
We compare our method to random sampling and a multi-resolution scan.
Results show that our proposed method outperforms both comparison methods, significantly reducing the number of sensing actions required for accurately mapping the embedded tumors.

\section{Related Work and Background}
A variety of methods have been developed to automate several tasks in robotic surgery. 
Jansen \textit{et al.}~\cite{jansen2009surgical} developed an automated method of tissue retraction to grasp deformable objects using a spring model. 
Elek \textit{et al.}~\cite{elek2017towards} automated blunt dissection of tissue using motion primitives. 
McKinley \textit{et al.}~\cite{mckinley2016interchangeable} developed an interchangeable surgical tool system to automate a multi-step tumor resection including palpation, incision, debridement, and injection. 
In this paper, we focus on autonomous sensing for tumor localization.

Automated sensing for tumor localization and segmentation has also been studied via palpation. 
Garg \textit{et al.}~\cite{garg2016tumor} presented an algorithm that samples over a stiffness map for localizing tumor boundaries with a palpation probe. 
Nichols \textit{et al.}~\cite{nichols2015methods} automated robotic palpation to localize and segment hard regions in soft tissues, also for tumor localization.
These methods are specific to palpation, framing the problem in two dimensions.
In this work, we consider a 3D sensor, such as a swept ultrasound, and model the geometry in a 3D continuous space.

Global optimization can enable autonomous surgical systems to optimize throughout a given search space by finding the global extremum of a given function. 
Bayesian optimization~\cite{Pelikan1999_GECCO} is one popular global optimization method and has been widely used in applications such as object surface estimation~\cite{yi2016active} and hyper-parameter optimization~\cite{bergstra2011algorithms}.
Bayesian optimization has also been used in surgical robotics, for instance as the optimization method in the above mentioned palpation work of Garg \textit{et al.}~\cite{garg2016tumor}.
Particularly effective in cases where function evaluation is computationally expensive, Bayesian optimization decides which points in the search space should be sampled and evaluated next via acquisition functions, such as expected improvement (EI)~\cite{jones1998efficient}. 
In this work, we leverage Bayesian optimization to determine our sensing poses, choosing an acquisition function that considers the balance between exploration and exploitation. 
Since autonomous surgical systems may have limited or inaccurate prior information regarding the number and size of tumors present in a given anatomical environment, it is necessary to simultaneously explore areas of high uncertainty as well as exploit existing knowledge of areas where tumors have already been sensed.

A variety of methods have been used to model a probabilistic distribution over a robot's environment.
For instance, Gaussian processes~\cite{rasmussen2003gaussian} have been used with Bayesian optimization as a standard modeling method.
Garg \textit{et al.}~\cite{garg2016tumor}, mentioned above, build a probabilistic model of the tissue stiffness map using Gaussian processes.
There have been other methods, e.g., ~\cite{o2012gaussian}, which model the probabilistic occupancy state of an environment using Gaussian Processes.
Senanayake \textit{et al.}~\cite{senanayake2017bayesian} developed Bayesian Hilbert maps to build an occupancy map in dynamic environments. 
They simultaneously introduced an extended version of Bayesian Hilbert maps, sequential Bayesian Hilbert maps~\cite{senanayake2017bayesian}, as a fast, sequential long-term occupancy mapping method in dynamic environments. 
In this work, we leverage sequential Bayesian Hilbert maps to build a probabilistic occupancy map updated sequentially via sensing.

\section{Problem Formulation}

In this work we assume that the sensor is noiseless, that segmentation in a sensed volume is perfect, and that the time required to perform the sensing action and associated segmentation dominates the time required to move the sensor between sensing poses.
We also assume that anatomy is rigid such that there is no deformation during sensing.

\begin{figure}[t!]
    \centering
	\includegraphics[width=\columnwidth]{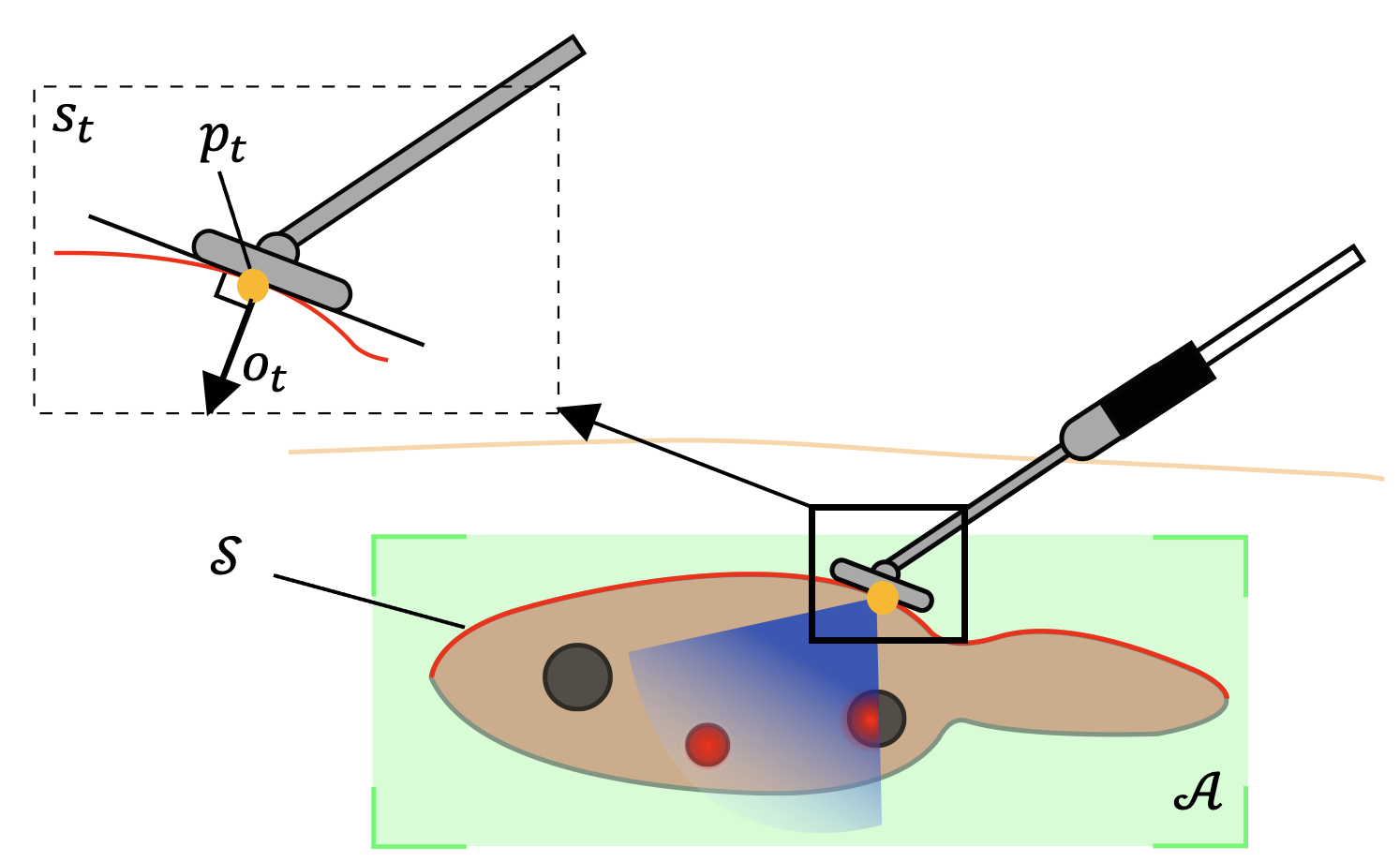}
    \caption{Example of the sensor pose at iteration $t$. The sensing pose $\vec{s}_{t}$ is composed of position $\vec{p}_{t}$ (yellow dot) and orientation $\vec{o}_{t}$ (black arrow)}
  \label{fig:problem}
\vspace{-12pt}
\end{figure}

We consider a case where tumors are embedded in an anatomical environment $\mathcal{A} \subset \mathbb{R}^{3}$, e.g., a patient's organ.
We define the tumors as $\mathcal{T} \subset \mathcal{A}$, a possibly disconnected set of arbitrary geometry.
We define a sensor workspace $\mathcal{S}$ that includes $N$ possible sensing poses $\vec{s}_{i}$ for $i = {1, \cdots, N}$.
A sensing pose $\vec{s}_{i}$ is a vector concatenating position and orientation, i.e., $\vec{s}_{i} = [\vec{p}_{i}, \vec{o}_{i}]$, where $\vec{p}_{i} \in \Reals^{3}$ and $\vec{o}_{i} \in SO(3)$ are the sensing position and orientation, respectively (see Fig. \ref{fig:problem}).
We define a general sensor model as the set of points, $v_i \subset \Reals^3$ that are sensed during a sensing action performed at $\vec{s}_{i}$, e.g., the volume visualized by an ultrasound sensing action performed when the transducer is centered and oriented at $\vec{s}_{i}$.

Let $\mathcal{S}_{\mathcal{M}} \subset \mathcal{S}$ then be an ordered sequence of $k$ sensing actions where $\mathcal{S}_{\mathcal{M}} = \{\vec{s}_{1}, \cdots, \vec{s}_{t}, \cdots, \vec{s}_{k} \}$ and $|\mathcal{S}_{\mathcal{M}}| = k \leq N$.
The sensed volume of a given sequence $\mathcal{S}_{\mathcal{M}}$ is then
\[
\mathcal{V}_{\mathcal{S}_{\mathcal{M}}} = 
\bigcup_{i=1}^{k} v_{i}.
\]

The goal then is to determine a sequence of sensing actions $\mathcal{S}_\mathcal{M}$ of minimal length (i.e., $|\mathcal{S}_{\mathcal{M}}|$), such that the geometry of all the tumors are mapped with high certainty.

\section{Method}
At a high level, our method is composed of two main pieces: (i) a 3D probabilistic occupancy map, implemented as a posterior likelihood distribution representing the likelihood that any point in space is occupied by a tumor, and (ii) an iterative optimization-based framework that reasons over the current distribution, determines the next sensing location, performs the sensing, and updates the distribution based on what was sensed.
Our method performs (ii) in a loop, iteratively updating (i) to improve the occupancy map, localizing the tumor(s) quickly and accurately.
The method is outlined in Algorithm~\ref{Alg: task_planner}.

\begin{algorithm}[t!]
\SetAlgoLined
\KwInput{Sensor workspace $\mathcal{S}$, Search space $\mathcal{A}$}
\KwOutput{A sequence of sensing configurations  $\mathcal{S}_{\mathcal{M}}$}
 initialize $\omega \leftarrow \mu_{0}, \Sigma_{0}$ \\ 
 $t \leftarrow 0$\\
  \While{time remains}{
  $t \leftarrow t+1$\\
  \eIf{t = 1}{
   $\vec{s}_{t}$ $\leftarrow$ random($\mathcal{S}$)\\
   $\mathcal{S}_{\mathcal{M}} \leftarrow \vec{s}_{t}$ 
   }{
   $\vec{s}_{t}$ $\leftarrow$ \texttt{NextQuery}($t$, $\omega$, $\mathcal{S}$, $\vec{x}_{*}$)\\
   $\mathcal{S}_{\mathcal{M}} \leftarrow$ \texttt{concatenate}($\mathcal{S}_{\mathcal{M}}, \vec{s}_{t}$) 
  }
  Acquire sensor data $\mathcal{D}_{t}$ given $\vec{s}_{t}$ \\
  $\mu_{t}, \Sigma_{t} \leftarrow$ \texttt{learn$\_$parameters}($\mathcal{D}_{t}$, $\omega$) \\
  $\omega \leftarrow \mu_{t}, \Sigma_{t}$ 
 }
 \textbf{\KwRet $\mathcal{S}_{\mathcal{M}}$}
 \caption{Sensor Sequence Planning}
 \label{Alg: task_planner}
\end{algorithm}

\begin{algorithm}[t!]
\SetAlgoLined
\KwInput{$t$, $\omega$, $\mathcal{S}$, $\vec{x}_{*}$}
\KwOutput{Next sensing pose  $\vec{s}_{t}$}
$\vec{a}_{t}$ $\leftarrow$ $\argmax_{\vec{x}_{*}} EI(\vec{x}_{*}, \omega)$ \\ %
$\vec{s}_{t}$ $\leftarrow$ \texttt{next$\_$sensing$\_$pose}($a_{t}$, $\mathcal{S}$)\\
\KwRet $\vec{s}_{t}$
 \caption{NextQuery}
 \label{Alg: next_query}
 \end{algorithm}

\subsection{Sensor Model}
\begin{figure}[t]
  \begin{center}
      \includegraphics[width=\columnwidth]{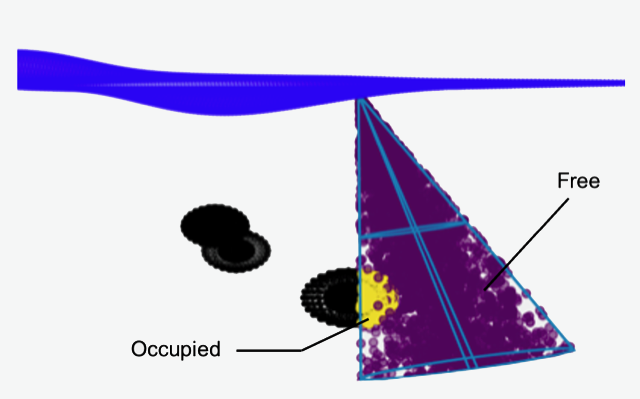}
	\caption{Example sensing action. In the cone-shaped sensing volume, oriented normal to the surface, the unoccupied free points (purple) and occupied tumor points (yellow) are determined.}
	\label{fig:data}
  \end{center}
  \vspace{-1em}
\end{figure}

In this work we consider a sensing action at sensing pose $\vec{s}_i \in \mathcal{S}$ to be a cone-shaped volumetric occupancy map of the anatomy in the cone.
In practice this could come from automated segmentation of a localized ultrasound sensing action, for instance.
This cone then defines the sensor volume $v_{i}$ (see Fig.~\ref{fig:data}).
Points within the cone are then labeled as either occupied (i.e., part of tumor geometry), or unoccupied (i.e., not tumor).
More formally, at iteration $t$, for each point in the cone $\vec{x} \in v_{t}$ we define an occupancy indicator $y \in \{0, 1\}$ denoting whether $\vec{x}$ is sensed as part of a tumor.
This then defines the tuple $(\vec{x}, y)$ for each point in $v_{t}$.
We define the set of these tuples for a given sensing action as the sensor data $\mathcal{D}_t$.
The sensor data $\mathcal{D}_t$ acquired from each measurement is used to update the sequential Bayesian Hilbert map parameters which define a probabilistic occupancy map of the unsensed anatomy, described below. 

\subsection{3D Occupancy Mapping}
We leverage sequential Bayesian Hilbert maps to model the occupancy states of given anatomical environments in an iterative manner.
Essentially, sequential Bayesian Hilbert maps define a classifier that estimates the probability of an unsensed point \(\vec{x}\) being occupied. 
Kernel functions define the sequential Bayesian Hilbert map features where for some kernel \(k\), $k(\vec{x}, \widetilde{\vec{x}}_{j})$,  the kernel evaluates the similarity between the query point \(\vec{x}\)  and a hinge point $\widetilde{\vec{x}}_{j}$ that is fixed at some location in the search space to be mapped. 
The feature vector $\Psi(\vec{x})$ represents the vector of kernel evaluations to all hinge points in the space, $\Psi(\vec{x}) = (k(\vec{x}, \widetilde{\vec{x}}_{1}), k(\vec{x}, \widetilde{\vec{x}}_{2}), \cdots)$.
Following~\cite{senanayake2017bayesian}, we fix $M$ hinge points $\widetilde{\vec{x}}_{j}$ for $j = {1, \cdots, M}$ spatially in $\mathcal{A}$ to compute the feature vector $\Psi(\vec{x}) \in \mathbb{R}^{1 \times M}$. 
We then define the likelihood of occupancy using the feature vector via a parametric logistic-regression model,
\[
    P(y|\vec{x},\vec{w}) = \sigma(\vec{w}\Psi^{T}(\mathbf{\vec{x}})),
\]
where $\sigma(\cdot)$ is the sigmoid function and $\vec{w} \in \mathbb{R}^{1 \times M}$ is a linear weight vector. 
After sensing observation $t$ of data collection, we model a normal distribution over the weight vector, $\vec{w} \sim \mathcal{N}(\mu_t, \Sigma_t)$, with mean $\mu_t \in \mathbb{R}^{1 \times M}$ and variance $\Sigma_t \in \mathbb{R}^{1 \times M}$. We define the vector of all parameters \(\vec{\omega} = \{\mu, \Sigma\}\). 
Note that in sequential Bayesian Hilbert maps the weight vector $\vec{w}$ is typically initialized with zero mean and high variance, representing that we do not have prior knowledge on the distribution. We use the squared exponential kernel in our implementation
\begin{equation}
k(\vec{x},\widetilde{\vec{x}}) = \mathtt{exp}\left(-\gamma||\vec{x}-\widetilde{\vec{x}}||^2\right)
\label{eq:kernel}
\end{equation}
where \(\gamma\) is a hyper-parameter defining the length scale of the kernel.

Given the model we then wish to learn the mean and variance of the parameter distribution $\omega$.
As its name suggests, sequential Bayesian Hilbert maps use Bayes' theorem to model the posterior distribution of the parameter $\omega$ as:
\[
    P(\vec{w}|\vec{x}, y) = \frac{P(y|\vec{x},\vec{w})P(\vec{w})}{P(y)}.
\]
The posterior over weights $P(\vec{w}|\vec{x}, y)$ cannot be explicitly computed  because of the combination of the sigmoidal likelihood and Gaussian prior, \cite{senanayake2017bayesian} provides a way to approximate the posterior $Q(\vec{\omega})$ defined also as Gaussians by estimating the parameters $\vec{\omega}$ through expectation-maximization (EM). 
The $\texttt{learn\_parameter($\cdot$)}$ function in Algorithm~\ref{Alg: task_planner} is a function for estimating the parameters $\vec{\omega}$. (See \cite{senanayake2017bayesian} for further details).

This then defines a continuous probabilistic occupancy map, e.g., for any query point $\vec{x}_{*} \in \mathcal{A}$, the probability of occupancy is defined as $P(y|\vec{x}_{*},\vec{w})$.

\subsection{Iterative Optimization-Based Framework}
Given the sequential Bayesian Hilbert map at a given iteration, we must determine the next sensing pose to gather the most relevant information and update the map.
This process is outlined in Algorithm~\ref{Alg: next_query}.
The sequential Bayesian Hilbert map provides a probabilistic occupancy map defined over the 3D anatomy $\mathcal{A}$, but the sequential Bayesian Hilbert map has no knowledge of our sensor workspace $\mathcal{S}$.
As such, we will first determine a point in $\mathcal{A}$ from the sequential Bayesian Hilbert map that should be sensed next, and then determine a pose in $\mathcal{S}$ that will do so effectively.

We leverage Bayesian optimization to determine the next query point $\vec{a}_{t} \in \mathcal{A}$. 
To do so, we must define an acquisition function based on the posterior distribution provided by the sequential Bayesian Hilbert map at the current iteration.
As is frequently the case in Bayesian optimization, in our problem it is important to define an acquisition function that balances exploitation and exploration.
Insufficient exploitation may lead to failure to fully map the tumors while insufficient exploration may result in not mapping tumors in unexplored regions. 
To balance these, we choose Expected Improvement (EI)~\cite{jones1998efficient} as our acquisition function.

More formally, EI is defined as
\[
    EI(\vec{x}_{*}) = \mathbb{E} [\max(0, f(\vec{x}_{*}) - f(\vec{x}^{+}))],
\]
where $f(\cdot)$ is an objective function which in our method is the occupancy likelihood distribution, $P(y|\vec{x}_{*},\vec{w})$, defined by the Bayesian Hilbert map at the current iteration,
and $f(\vec{x}^{+})$ is the highest value of the distribution.
The expected improvement can be expressed in closed form \cite{jones1998efficient}:
\begin{align}
\begin{split}
    EI(\vec{x}_{*}, \omega) = (\mu - f(\vec{x}^{+}) - \xi)\Phi(\frac{\mu - f(\vec{x}^{+}) - \xi}{\sigma}) \\
    + \Sigma \phi(\frac{\mu - f(\vec{x}^{+}) - \xi}{\sigma})
\end{split}
\label{eq:aquisition}
\end{align}
where $\Phi(\cdot)$ is the cumulative distribution function, $\phi(\cdot)$ is the probability density function, and $\xi$ is an exploration parameter.
Mean $\mu$ and standard deviation $\Sigma$ are also computed via the sequential Bayesian Hilbert map. 
We are able to determine the next query point $\vec{a}_{t} \in \mathcal{A}$ by optimizing the acquisition function, $\vec{a}_{t} = \argmax_{\vec{x}_{*}}EI(\vec{x}_{*}, \omega)$. 

\begin{figure}[t]
  \begin{center}
      \includegraphics[width=\columnwidth]{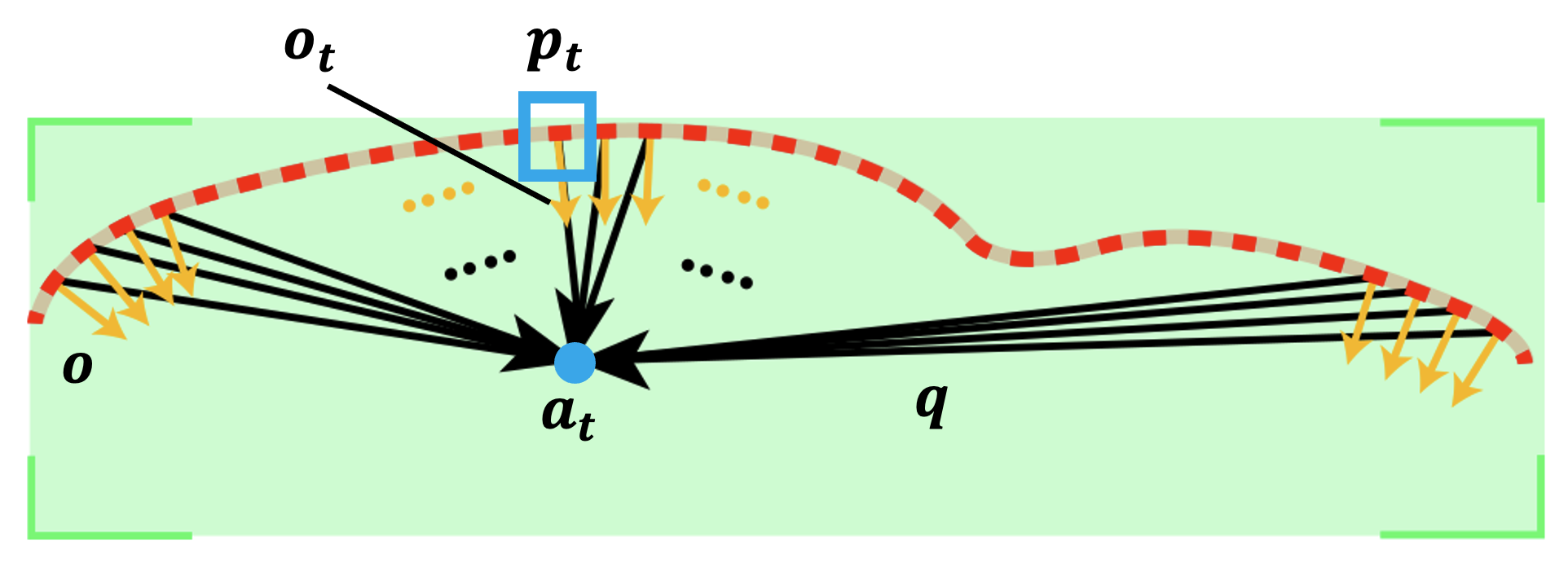}
	\caption{Determining the next sensing pose $\vec{s}_t = (\vec{p}_{t}, \vec{o}_t)$. When a query vector (black vectors) and a sensing pose's surface normal (orange vectors) are aligned (blue box), the corresponding pose is selected.}
	\label{fig:query}
  \end{center}
  \vspace{-1.5em}
\end{figure}

We next must determine the sensing pose $\vec{s}_{t}$ corresponding to the query point $\vec{a}_{t}$ ($\texttt{next$\_$sensing$\_$pose}$ in Algorithm~\ref{Alg: next_query}, line 2).
To do so, we choose a sensing pose that has an orientation that closely aligns with the vector defined by the pose's position and the query point, i.e., a pose that points toward the query point (see Fig.~\ref{fig:query}).
We search over the sensing poses and for each define a query vector $\vec{q}$ as the vector between the given sensing pose's position $\vec{p}$ and the query point $\vec{a}_t$.
We then determine the angular difference between the sensing pose's orientation $\vec{o}$ and $\vec{q}$.
We select the first sensing pose found such that this angle is below a given threshold.

\subsection{Combined Method}
Combining the pieces above we get the full method, outlined in Algorithm~\ref{Alg: task_planner} and Algorithm~\ref{Alg: next_query}.
The method takes as input the sensor workspace $\mathcal{S}$ and the region to search over $\mathcal{A}$, and outputs a sequence of sensing configurations $\mathcal{S}_{\mathcal{M}}$ to detect the tumor(s) $\mathcal{T}$ embedded in the anatomical environment.
We first initialize the parameter $\omega$ of each kernel of the Bayesian Hilbert maps (Algorithm~\ref{Alg: task_planner} line $1$).
The first sensing pose $\vec{s}_{1}$ is then randomly chosen from $\mathcal{S}$ as we do not yet have knowledge of the distribution (Algorithm~\ref{Alg: task_planner} line $6$). 
Afterwards, sensing poses are determined by Algorithm~\ref{Alg: next_query} (called by line $9$ of Algorithm~\ref{Alg: task_planner}) and the sequence of poses is accumulated in $\mathcal{S}_{\mathcal{M}}$ ($\texttt{concatenate}$ in Algorithm~\ref{Alg: task_planner}, line $10$). 
For each sensing pose, the corresponding sensor data $\mathcal{D}_t$ is collected (Algorithm~\ref{Alg: task_planner} line $12$) and is then used to update the posterior distribution of the parameter $\omega$ (Algorithm~\ref{Alg: task_planner} line $13$). 
The method repeats as time allows, refining the estimation of the tumor volumes with increasing iterations.

\begin{figure*}[t]
    \centering
    \includegraphics[width=\linewidth]{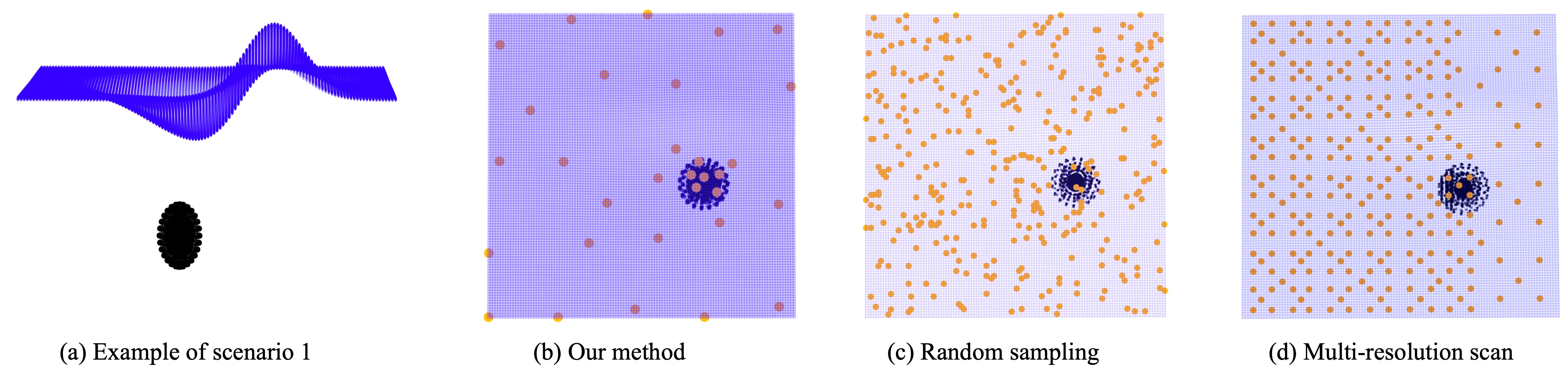}
  \caption{Scenario 1, trial 1. The dots (orange) indicate the sensing poses chosen by the methods to identify the embedded tumor. (a) The surface (blue) and tumor (black). (b) Our method requires 31 samples while balancing the exploration of the entire search space and the exploitation of regions where tumors may lie. (c) Random sampling requires 558 samples. (d) Multi-resolution scan requires 270 samples.}
  \label{fig:scene1_result}
\end{figure*}

\section{Experiments and Results}
We evaluate our method in two ways. 
First, we evaluate the method's efficiency and compare against random sampling and a multi-resolution scan as strategies for determining sensing pose sequences.
We do so in synthetic, randomly generated example scenarios and demonstrate our method is capable of localizing the embedded tumors with significantly fewer sensing actions.
Second, we evaluate our method's accuracy as a function of the number of sensing actions in the presence of multiple tumors.
We do so both in a synthetic environment with randomly placed tumors and in a real medical scenario segmented from a patient CT scan in the liver tumor segmentation (LiTS) dataset \cite{bilic2019liver}. 
For all experiments we set the parameters $\gamma = 5$ in the kernel function~(\ref{eq:kernel}), $\xi = 0.01$ in the acquisition function~(\ref{eq:aquisition}), and use $2890$ hinge points distributed on a 3D grid in the environments.

\subsection{Evaluating Efficiency}
To evaluate our method's sampling efficiency quantitatively, we generate a synthetic organ surface and place a spherical tumor volume randomly below it (see Fig.~\ref{fig:scene1_result} (a)).
We refer to this as scenario 1.
The randomly placed sphere-shaped tumor is composed of $500$ data points and the sensor workspace is composed of 14,400 data points.
The synthetic surface is made uneven to approximate the uneven nature of the surface of human organs and is generated via the $\texttt{get\_test\_data}$ function provided by Matplotlib, a graphical plotting library for Python.
We define the positions in $\mathcal{S}$ as the points on the surface and compute the orientation of each sensing pose as the surface normal perpendicular to a tangent plane fit to each surface point's 10 nearest neighbors.

We compare our method against random sampling, in which we draw sample poses uniformly at random from $\mathcal{S}$; and a coarse-to-fine multi-resolution scan, in which the search space is divided into increasingly fine cells in each round, i.e., in the first round there is one cell containing the entire search space, in the second round the space is divided into four cells, etc.
At each round we apply sensing actions to the center of each cell, ordered as a raster scan.
We measure the number of samples required by each method to detect the tumor, defined as having sensed $95\%$ of the tumor points.

\begin{figure}[t]
  \begin{center}
      \includegraphics[width=\columnwidth]{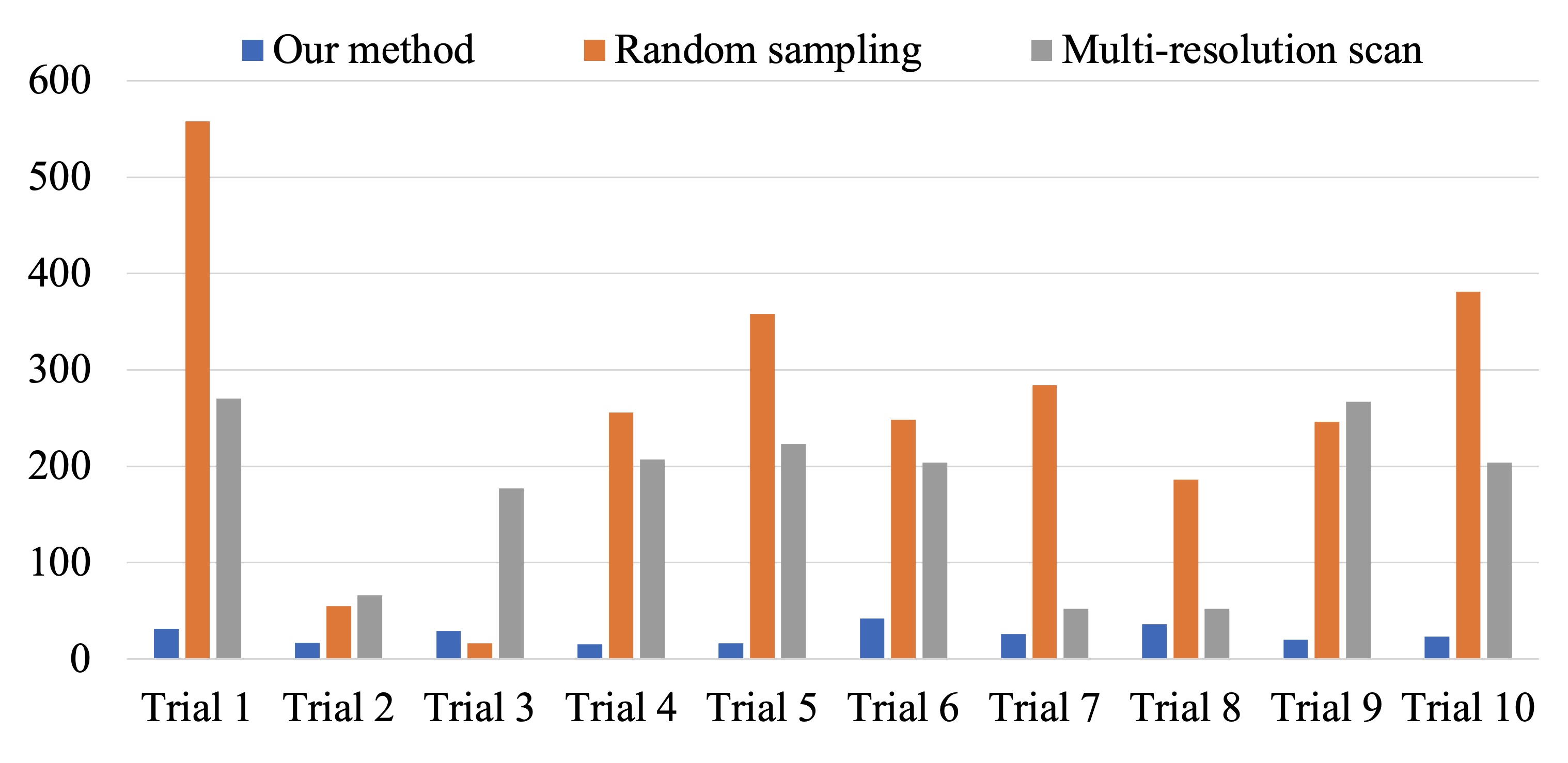}
	\caption{The number of sensing pose samples required by each method for 10 random trials. The mean and standard deviation across the trials are $25.5\pm 9.0$ for our method, $258.8 \pm 156.7$ for random sampling, and $172.2 \pm 84.7$ for the multi-resolution scan.
	}
	\label{fig:comparison}
  \end{center}
  \vspace{-2.0em}
\end{figure}

We average the results over ten trials in which the tumor is randomly placed below the surface.
We demonstrate the results for all methods across all trials in Fig.~\ref{fig:comparison}.
As can be seen, our method detects the tumor with much fewer samples.
Compared to our method, the random sampling method required on average approximately $10.1$ times the number of sensing actions ($258.6$ required by random sampling compared with $25.5$ required by our method) and the multi-resolution scan required approximately $6.5$ times the number of sensing actions ($172.2$ required by the multi-resolution scan compared with $25.5$ required by our method) to fully sense the tumor.
The specific sensing poses determined by the method for one of the trials are shown in Fig.~\ref{fig:scene1_result}.

This analysis demonstrates our method's ability to efficiently generate sensor poses that sense the tumor geometry.

\subsection{Evaluating Accuracy}

\begin{figure*}[t]
    \captionsetup{skip=-1pt}
    \centering
    \includegraphics[width=\linewidth]{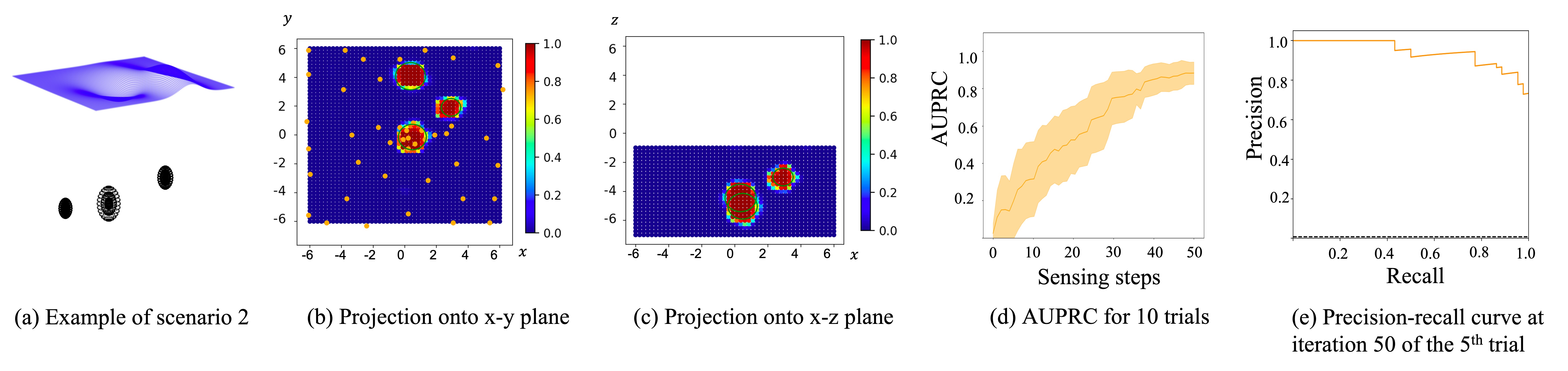}
  \caption{(a) Scenario 2 consists of three sphere-shaped tumors and the same sensor workspace as scenario 1. The tumors are randomly positioned in each trial. (b) Top view (x-y plane) of the final occupancy map for the $5^{th}$ trial. The orange dots indicate the sensing poses chosen by our method. (c) The side view (x-z plane) of the occupancy map. (d) The mean (orange line) and standard deviation (shaded area) of the AUPRC as the iterations increase, averaged across 10 trials. The AUPRC converges to $0.88$ within 50 sensing iterations for all trials. (e) The precision-recall curve at the last iteration of the $5^{th}$ trial, with AUPRC of $0.89$. }
  \label{fig:scene2_results}
\vspace{-12pt}
\end{figure*}

\begin{figure*}[t]
    \captionsetup{skip=-1pt}
    \centering
    \includegraphics[width=\linewidth]{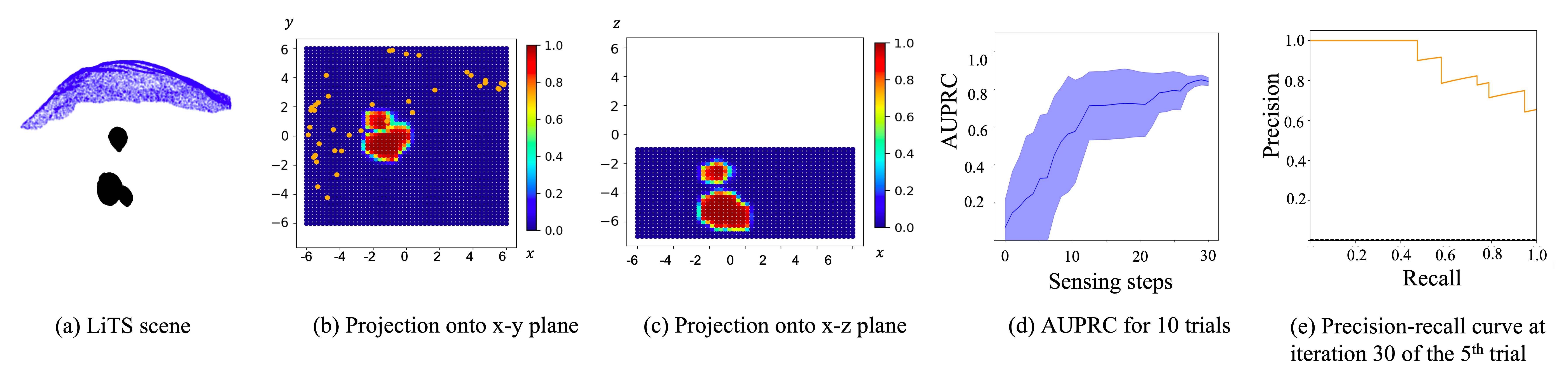}
  \caption{(a) The LiTS scenario is composed of three tumors in the liver. (b) Top view of the final occupancy map of the $5^{th}$ trial. The orange dots indicate the sensing poses chosen by our method. (c) The side view of the occupancy map. (d) The mean (blue line) and standard deviation (shaded area) of the AUPRC as the iterations increase, averaged across 10 trials. The AUPRC graph converges to $0.84$ within 30 steps for all trials. (e) The precision-recall curve at the last iteration of the $5^{th}$ trial, with AUPRC of $0.84$.}
  \label{fig:LiTS_results}
 \vspace{-20pt}
\end{figure*}

\begin{figure}[t]
    \centering
   \begin{subfigure}[b]{0.23\textwidth}
	\includegraphics[width=\textwidth]{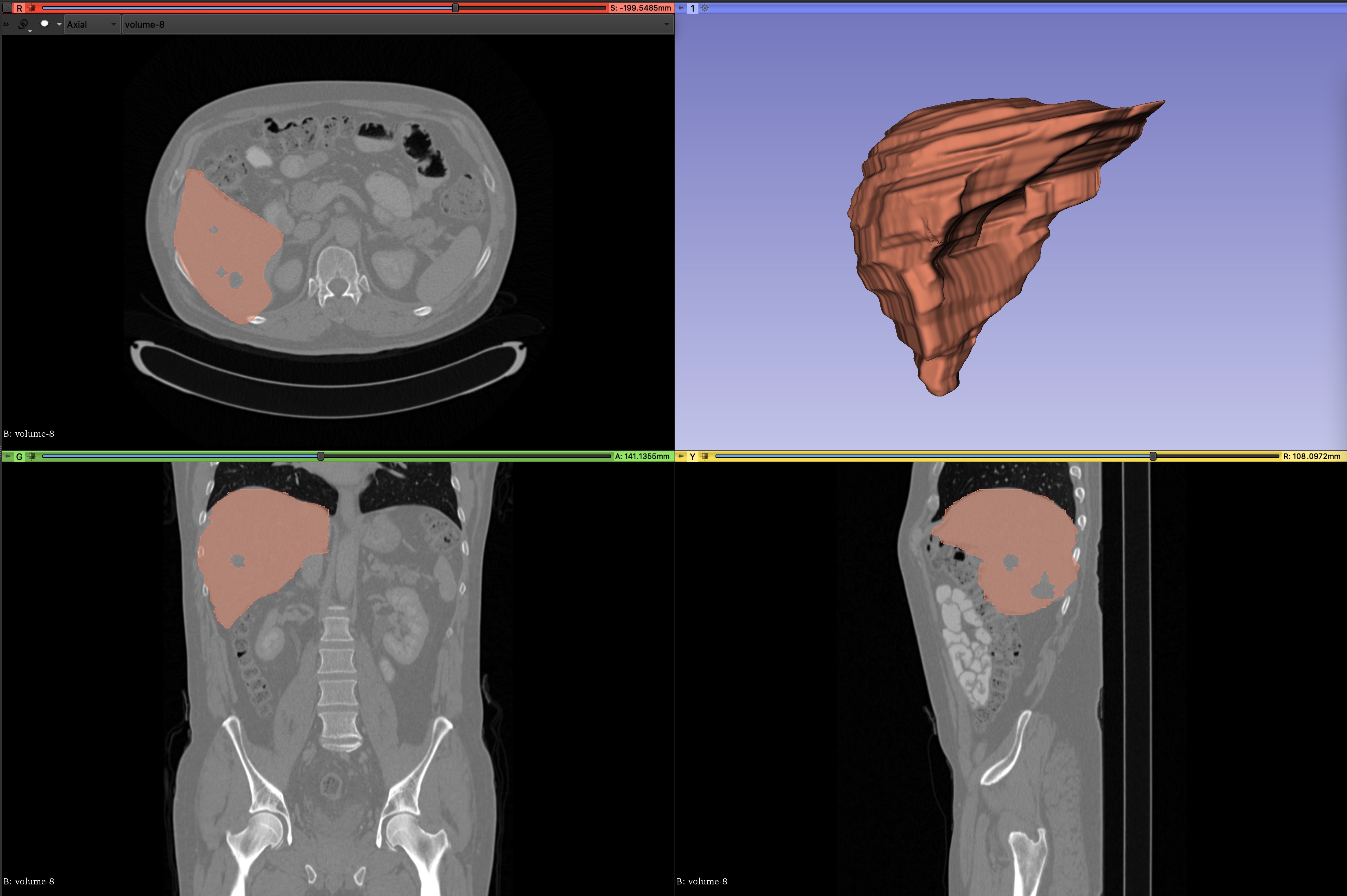}
    \caption{CT scan and segmentation}
    \label{fig:ct_scan}
  \end{subfigure}
  \begin{subfigure}[b]{0.23\textwidth}
  \captionsetup{skip=-1pt}
	\includegraphics[width=\textwidth]{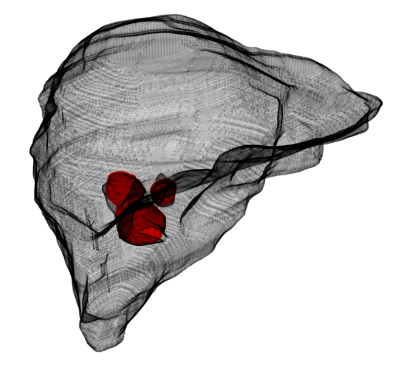}
    \caption{3D rendering}
    \label{fig:LiTS_seg}
  \end{subfigure}
  \caption{Liver and tumors segmented from the LiTS data. (a) The liver and tumors were segmented via 3D Slicer~\cite{fedorov20123d}. (b) The resulting 3D geometry utilized for evaluation.}
  \label{fig:LiTS}
\vspace{-1.5em}
\end{figure}

A notable property of the Bayesian Hilbert occupancy map is that it incorporates probabilistic information about regions not yet sensed, which is refined with more sensing actions.
Here we evaluate the accuracy of the occupancy map generated by our method as the number of sensing actions performed increases.
We consider two scenarios, a synthetic scenario with three tumors randomly placed, named scenario 2 (see Fig~\ref{fig:scene2_results} (a)), and a real-life scenario with multiple tumors in a patient's liver, segmented via 3D Slicer~\cite{fedorov20123d} from a CT scan in the liver tumor segmentation (LiTS) dataset \cite{bilic2019liver}, which we name the LiTS scenario (see Fig.~\ref{fig:LiTS_results} (a) and Fig.~\ref{fig:LiTS} (b)).
$\mathcal{S}$ for scenario 2 is generated as for scenario 1, and $\mathcal{S}$ for the LiTS scenario is generated by taking points on the top surface of the liver segmentation (see Fig.~\ref{fig:LiTS_results} (a)) and defining surface normals as in scenarios 1 and 2.

We note that the probabilistic occupancy map being produced by our method acts as a binary classifier when combined with a threshold value.
As a metric to evaluate binary classifiers, both the area under the receiver operating characteristics (AUROC) curve and the area under the precision-recall curve (AUPRC) are popular metrics. 
However, the ROC tends to provide an optimistic view of the performance when it comes to imbalanced datasets, potentially resulting in incorrect interpretation~\cite{saito2015precision}. 
As our case is highly imbalanced, e.g., approximately a 200:1 ratio of free space to tumor volume in scenario 2 and a 300:1 ratio in the LiTS scenario, we choose the AUPRC as our evaluation metric. 

For both scenario 2 and the LiTS scenario, we evaluate the AUPRC at each iteration, i.e., sensing action, and evaluate how the AUPRC improves as the iterations increase.
At each iteration the Bayesian Hilbert occupancy map is refined, becoming more accurate, and the AUPRC improves.
As our method is subject to random initialization, we average the results across 10 random trials with different initializations.

Fig. \ref{fig:scene2_results} shows the results for scenario 2.
To evaluate correct classification we discretize the search space into 6292 points on a 3D grid.
Fig. \ref{fig:scene2_results} (d) shows the mean and standard deviation of the AUPRC for 10 trials as the sensing iterations increase, converging to $0.88$ at 50 iterations. 
A high AUPRC indicates that the method correctly labels the positive examples (tumors) without falsely labeling the negative examples (free space) as positive.
Considering that we have a highly imbalanced environment, the baseline of an uninformed model would be $0.006$ (the ratio of positives to negatives).
The resulting AUPRC of $0.88$ indicates high accuracy.
Fig.~\ref{fig:scene2_results} (e) shows the specific precision-recall curve at iteration 50 for one of the trials, and Fig.~\ref{fig:scene2_results} (b) and (c) show two views of the occupancy map.

Fig. \ref{fig:LiTS_results} shows the results for the LiTS scenario.
We task the method with correctly classifying the tumors inside the liver (using the tumor segmentations as ground truth).
In this scenario we discretize the liver search space into 6292 evenly spaced points on a 3D grid for evaluation.
As with scenario 2, we average over 10 runs.
In the LiTS scenario our method converges to an average of $0.84$ AUPRC after 30 sensing iterations, demonstrating high accuracy when compared with the baseline of $0.003$ of an uninformed model for this scenario.

\section{Conclusion}
In this work, we presented a method for planning sensing actions that models the anatomical environment using sequential Bayesian Hilbert maps and determines the sensing strategy using Bayesian optimization.
We evaluated our method in three anatomical scenarios, a synthetic scenario with a single tumor, a synthetic scenario with multiple tumors, and a real-life case of tumors in a patient's liver segmented from a CT scan. 
In the first scenario, we evaluated the efficiency of our method by comparing it with random sampling and a multi-resolution scan, showing that it outperforms other methods with respect to the number of sensing actions required. 
In the other two scenarios, we demonstrated the ability of our method to accurately map the tumors.

In future work, we plan to relax our assumptions.
This includes augmenting our method with the ability to consider deformable organs and other anatomical features beyond tumors.
We also plan to integrate real sensing via, e.g., ultrasound and implement the method on a physical robot. 
Further, we plan to extend the method to consider the sensing path rather than just discrete sensing actions.

\bibliographystyle{IEEEtran}

\bibliography{reference}

\end{document}